\documentclass[%
  reprint,          
  superscriptaddress,
  amsmath,amssymb,  
  aps,              
]{revtex4-2}%

\usepackage[T1]{fontenc}
\usepackage{graphicx}
\usepackage{amsmath}

\begin{document}

\title{Rethinking the Illusion of Thinking}

\author{Iñaki Dellibarda Varela}
\email{Corresponding author: i.dellibarda@car.upm-csic.es}
\affiliation{Center for Automation and Robotics, Spanish National Research Council (CSIC-UPM), Madrid, Spain}

\author{Pablo Romero-Sorozabal}
\affiliation{Center for Automation and Robotics, Spanish National Research Council (CSIC-UPM), Madrid, Spain}

\author{Eduardo Rocon}
\affiliation{Center for Automation and Robotics, Spanish National Research Council (CSIC-UPM), Madrid, Spain}

\author{Manuel Cebrian}
\affiliation{Center for Automation and Robotics, Spanish National Research Council (CSIC-UPM), Madrid, Spain}

\date{\today}


\date{\today}          

\begin{abstract}
Earlier this year, Apple ignited controversy by publishing "The Illusion of Thinking," prompting heated debate within the AI community. Critics seized upon the findings as conclusive evidence that Large Reasoning Models (LRMs) lack genuine reasoning capabilities, branding them as mere stochastic parrots. Meanwhile, defenders—spearheaded by Lawsen et al. (2025)—fired back, condemning the experimental setup as flawed and the conclusions overstated. We clarify this debate by replicating and refining two of the original study’s most contentious benchmarks: Towers of Hanoi and River Crossing. By introducing incremental stepwise prompting and agentic collaborative dialogue, we show that previously reported failures solving the Towers of Hanoi were not purely result of output constraints, but also partly a result of cognition limitations: LRMs still stumble when complexity rises moderately (around 8 disks). Moreover, the River Crossing results initially heralded as catastrophic failures turn out to hinge upon testing unsolvable configurations. Once we limit tests strictly to solvable problems—LRMs effortlessly solve large instances involving over 100 agent pairs. Our findings ultimately defy simplistic narratives:  today’s LRMs are stochastic, RL-tuned searchers in a discrete state space we barely understand. Real progress in symbolic, long-horizon reasoning demands mapping that terrain through fine-grained ablations like those introduced here.
\end{abstract}

\maketitle

%


%
%
\section{Introduction}

As Large Reasoning Models (LRMs) advance, reliably assessing their performance has become correspondingly challenging. Benchmark leakage, prompt sensitivity, and high mis-grading rates on complex tasks obscure true capability. Consequently, refining evaluation protocols has become as critical to progress as advances in model architecture ~\cite{zhou2025general}.

In June 2025, Apple published a highly controversial paper, \textit{The Illusion of Thinking: Understanding the Strengths and Limitations of Reasoning Models via the Lens of Problem Complexity} by Shojaee et al.~\cite{illusion-of-thinking}. The authors evaluated the reasoning mechanisms and capabilities of frontier LRMs relying on controllable puzzles rather than standard benchmarks to systematically probe reasoning across varying levels of complexity.

Based on their study, the authors concluded that current LRMs exhibited fundamental limitations in generalizable reasoning. These models consistently failed to perform beyond a certain complexity threshold: they underperformed on high-complexity tasks and tended to overthink in simple ones, where older models—namely Large Language Models (LLMs)—often achieved better results. Their performance also varied significantly across puzzles, successfully executing hundreds of steps in the \textit{Towers of Hanoi} before failing, while rarely completing more than five moves in the \textit{River Crossing} problem.

Another notable finding concerned token usage. LRMs reach peak token consumption when the task is complex but still solvable. Conversely, when the model implicitly detects that the task lies beyond its capabilities, it drastically reduces its output length—suggesting as potential form of internal uncertainty estimation or early abandonment strategy.

The study rapidly attracts attention beyond technical circles and sparks media coverage in outlets such as \textit{The Guardian} and \textit{The Wall Street Journal}, which highlight its warnings about the collapse of LRMs in complex reasoning tasks and raise concerns about the scalability of AI reasoning~\cite{Guardian,WSJ}. In parallel, a number of rapid-response publications and technical commentaries emerge, critically analyzing the study's methodology, results, and conclusions~\cite{lawsen2025,khan2025}.

The study generated a sharp divide within the AI community. On one side, critics interpreted the findings as definitive evidence that LRMs fundamentally lack reasoning capabilities, reducing them to mere stochastic parrots. On the other, defenders argued that the study suffers from flawed benchmarks, questionable experimental design, and overreaching conclusions. The debate quickly transcended academic boundaries. This polarization highlights the urgency of reassessing how we evaluate reasoning in language models, and whether current benchmarks truly reflect cognitive competence or merely expose interface limitations.

Building on this debate, our work emphasizes the influence of task formulation, input validity, and prompt structure on LRM performance. Through our experimental reframing, we observed that structured methodologies—such as stepwise resolution and agentic interaction—can significantly improve results in certain conditions. However, we also identified persistent failure modes that reveal limitations in long-horizon consistency and symbolic generalization. Our analysis suggests that these reasoning breakdowns stem not only from architectural constraints, but also from the inherently stochastic nature of these systems and the optimization methods they rely on. This underscores the need for more nuanced evaluation protocols that go beyond surface-level success rates.


It is therefore critical to rigorously characterize these models’ limitations, capabilities, and behavioral patterns, and to establish robust evaluation frameworks that can assess their performance in ways aligned with the demands of scientific reasoning~\cite{omega2024}.

We emphasize that this work does not aim to undermine the contributions of Shojaee et al.~\cite{illusion-of-thinking}. On the contrary, we consider their study both impactful and timely. It opens a valuable discussion on the nature of reasoning in LRMs, and our goal is to complement their findings by introducing alternative perspectives and experimental refinements.

\section{Foundational Work}

\begin{figure}
    \centering
    \includegraphics[width=0.7\linewidth]{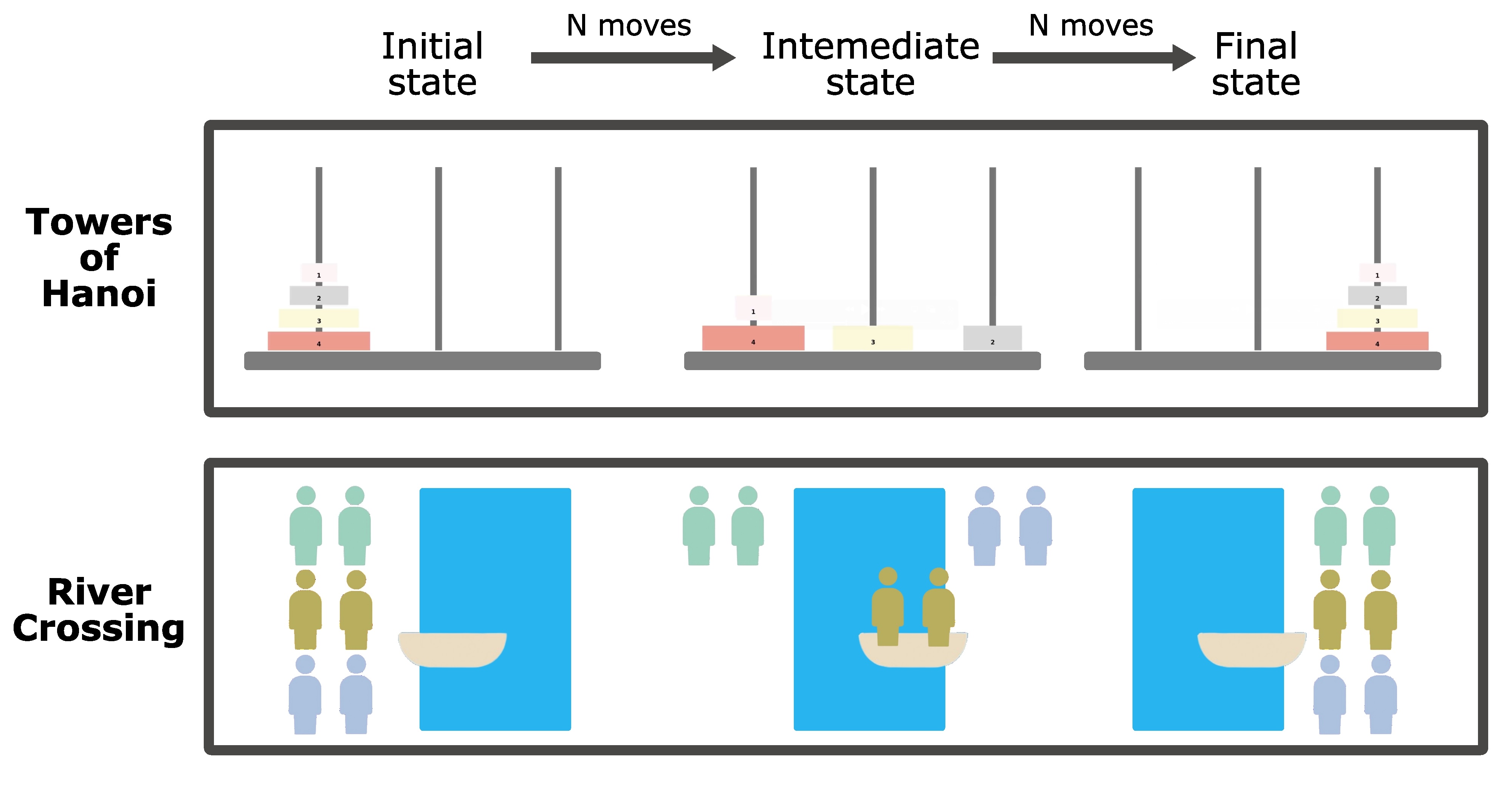}
    \caption{Illustration of the two benchmark puzzles analyzed in this study: \textit{Towers of Hanoi} (top) and \textit{River Crossing} (bottom). Each panel shows three sequential states: the initial configuration, an intermediate state after a few moves, and the final goal configuration. In Towers of Hanoi, the disks must be moved from the left peg to the right peg following size and move constraints. In River Crossing, each pair of agent and actor (same color) must cross the river while respecting the constraint that no actor can be left with a different agent unless accompanied by their own.}
    \label{fig:puzzles}
\end{figure}

Aforementioned, Shojaee et al.~\cite{illusion-of-thinking} proposed a set of puzzle-based benchmarks to evaluate the reasoning capabilities of Large Reasoning Models (LRMs). Among these, \textit{Towers of Hanoi} and \textit{River Crossing} (see Fig.~\ref{fig:puzzles}) emerged as the most discussed and revealing tasks in their study.

\textbf{Towers of Hanoi} is a classic recursive puzzle involving three pegs and a stack of disks of decreasing size placed on one peg in sorted order. The goal is to move the entire stack to a target peg, following two constraints: only one disk may be moved at a time, and no disk may be placed on top of a smaller one. The minimal number of moves required to solve the puzzle grows exponentially with the number of disks, following the expression $2^n - 1$. This makes it a scalable benchmark for evaluating long-horizon planning and symbolic rule-following. Shojaee et al.\ reported that LRMs handled the initial moves but deteriorated sharply beyond \(n\!\approx\!8\): models violated rules, repeated states, or produced invalid sequences. 

This trend was interpreted as a symptom of reasoning breakdowns in long-horizon planning. Although the minimal solution path was known and recursive, the models failed to generalize the underlying pattern, especially when prompted to generate full solutions in one pass. Interestingly, when the task exceeded the models’ operational limits, the number of tokens used dropped significantly, suggesting that the models implicitly recognized their inability to complete the solution and reduced their effort accordingly. Additionally, the authors found that in simpler configurations, traditional LLMs—without explicit reasoning chains—often outperformed LRMs, as they produced shorter, more direct outputs with fewer computational resources.

\textbf{River Crossing}. In this variant, $N$ agent-actor pairs must cross a river using a boat with limited capacity $k$, subject to the constraint that no actor may remain with a different agent unless its own agent is present. Difficulty grows with \(N\) and depends on \(k\); only instances satisfying \(k\!\ge\!4\) or \(N\!\le\!2k-1\) are solvable~\cite{efimova2018rivercrossingproblemsalgebraic,spahn2022variationsmissionariescannibalsproblem}. 

Empirically, in Shojaee et al.'s experiments, LRMs seldom produced more than five valid moves and showed the weakest results across the benchmark. However, these experiments also prompted the model to solve mathematically unsolvable initial configurations, highlighting critical limitations in the evaluation design.

These results were taken as evidence of a deeper limitation: although LRMs displayed emergent reasoning behaviors in some tasks, they lacked the robustness and consistency required to perform reliably across different problem domains.

\section{Experimental Design}
\subsection{Baseline Design}
Our experimental procedures build directly upon the baseline design proposed by Shojaee et al.~\cite{illusion-of-thinking}. To ensure compatibility, we reuse the original prompts (see Code Availability) provided in their study with only minor modifications tailored to our setting.

We employ the Gemini family of models as our LRM, specifically Gemini 2.5 Pro. This model demonstrates state-of-the-art performance in structured reasoning tasks~\cite{gemini2.5report,geminiteam2024,geminiteam2024geminifamilyhighlycapable}.



\subsection{Experimental Reframing Strategies}
We present alternative methodologies designed to address the limitations observed in the original benchmarks. Each subsection describes a distinct configuration combining task type with specific prompting strategies, aiming to uncover deeper insights into LRM reasoning performance.

\subsubsection{Stepwise Resolution in Towers of Hanoi:} The original publication shows that as the problem size increases—specifically, as the number of disks in the Towers of Hanoi grows—LRMs become increasingly unable to solve the task. Beyond a certain complexity threshold, models not only fail to generate valid solutions but also tend to produce shorter, incomplete outputs. This reduction in token usage suggests a form of implicit uncertainty management, where the model appears to allocate fewer resources when it “predicts” that the task is likely unsolvable or beyond its capabilities.

One widely held explanation for the LRM's failure in long-horizon tasks like Towers of Hanoi is not a lack of reasoning capacity, but rather a technical limitation. As the number of disks increases, the length of the output required to reach a valid solution grows exponentially, quickly approaching or exceeding the model’s maximum output window. Faced with this constraint, the model seems to reduce its effort, generating fewer tokens and often halting early. This behavior suggests that the model anticipates the impracticality of completing the task within the allowed token limit and, rather than wasting resources, implicitly chooses to abandon the attempt. These failures thus reflect constraints on solution length and output budgeting, rather than a fundamental inability to reason~\cite{lawsen2025}.

In order to investigate whether token budget limitations are the true cause of systematic failure in large puzzle instances, we adopt a stepwise resolution approach. The overall goal remains the same as in the original experiment, but instead of prompting the LRM to solve the entire problem in a single pass, we divide the task into $N$ subproblems. In each subproblem, the model is asked to generate the next $p$ steps toward the solution, starting from the current configuration. The subsequent prompt then resumes from the final state of the previous iteration. This iterative setup reduces the output burden at each stage, allowing us to test whether performance improves when the model operates under a shorter reasoning horizon.

\subsubsection{Agentic Dialogue in Towers of Hanoi:} Similar to the previous approach, this procedure also employs stepwise resolution. However, instead of prompting a single LRM with isolated queries to transition between states (separated by $p$ moves), we implement an agentic system composed of two LRMs that collaborate through dialogue. Each agent maintains access to a shared memory and infers the current state based on the latest movements provided by its counterpart. During each turn, the agent receives the last set of moves and is tasked with proposing the next $p$ steps toward the solution.

Recent studies have shown that multi-agent systems offer several advantages over single-model architectures, especially in tasks requiring symbolic reasoning, strategic planning, and coordination. Cooperative agents tend to explore more diverse solution paths, generate higher-quality answers through iterative refinement, and maintain more consistent long-horizon coherence than isolated LRMs ~\cite{sreedhar2024,qian2025,gao2025,wang2024,buehler2025r,crandall2018cooperating}. These findings support the use of agentic interaction not merely as a heuristic aid, but as a structured enhancement in solving recursively formulated problems like the Towers of Hanoi .

\subsubsection{River Crossing:} In the original study, this puzzle yields the poorest performance, with LRMs failing to solve any configuration where the number of agent–actor pairs $N>5$. However, the experimental setup introduces a critical flaw: while increasing $N$, the boat capacity $k$ remains fixed at 3. Formal solvability criteria for this class of problems establish that a configuration is solvable if and only if it satisfies one of the following conditions~\cite{efimova2018rivercrossingproblemsalgebraic,spahn2022variationsmissionariescannibalsproblem}:

\[
\begin{aligned}
&\text{(a)}\quad k \geq 4 \quad \Rightarrow\quad \forall N \geq 1, \text{ the puzzle is solvable},\\
&\text{(b)}\quad k \leq 3 \quad \Rightarrow\quad N \leq 2k - 1 \quad (\text{i.e., } N \leq 5 \text{ for } k=3).
\end{aligned}
\]

By fixing $k=3$ and increasing $N$ beyond 5, the original experiments include configurations that are unsolvable, invalidating the assessment of model reasoning capabilities in River Crossing tasks.

To ensure a fair evaluation, we replicate the original River Crossing experiments while restricting our tests to configurations that are mathematically solvable under the known constraints. This allows us to isolate reasoning failures from structural impossibility and to more accurately assess the LRM's capacity for symbolic rule-following under valid conditions.

\section{Results}
\subsection{Towers of Hanoi}
\subsubsection{Stepwise Resolution:}To evaluate the LRM's ability to solve the Towers of Hanoi puzzle under stepwise prompting, we test configurations of increasing difficulty by varying the number of disks from $N=3$ to $N=10$. For each configuration, we run 10 independent trials to obtain robust averages. In this setup, the full puzzle is decomposed into a sequence of substages, with the model instructed to generate a fixed number of $p$ moves toward the solution at each substage.

\begin{figure}[h]
    \centering
    \includegraphics[width=\linewidth]{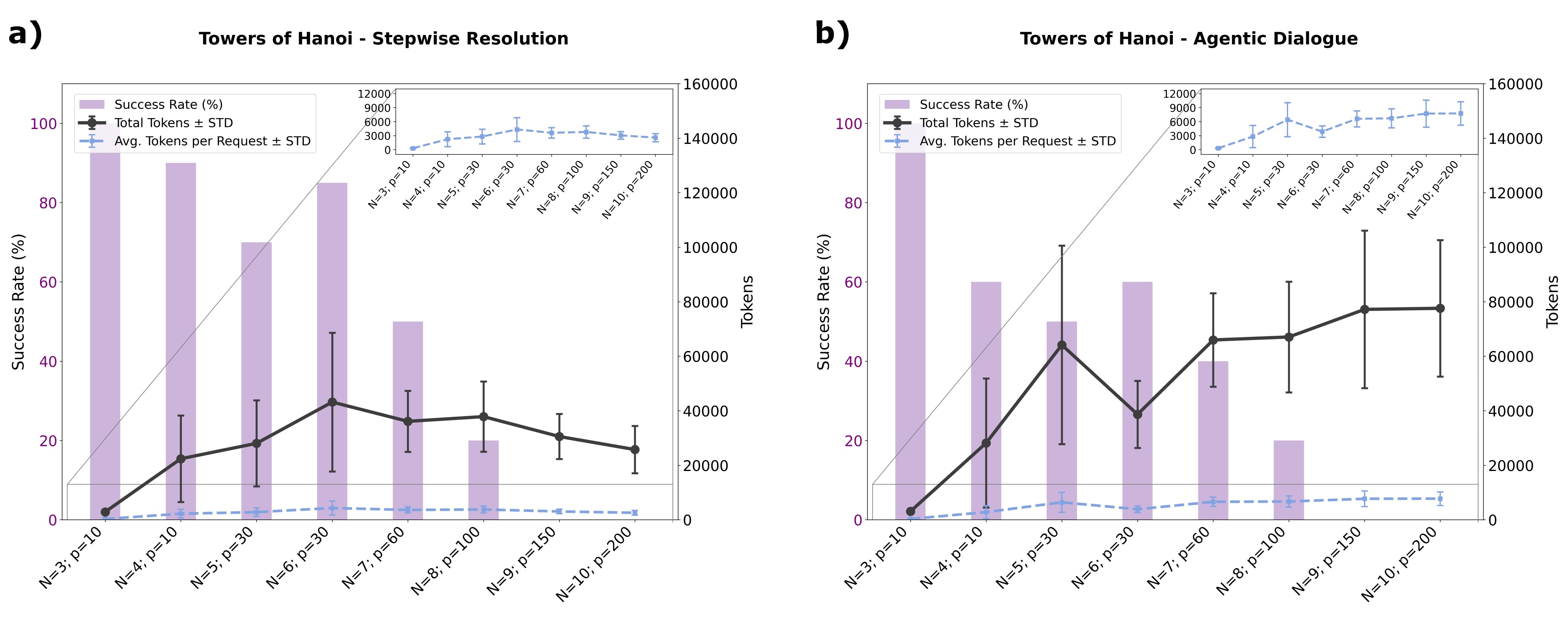}
    \caption{Comparison of performance across two solving strategies for the Towers of Hanoi puzzle: \textbf{(a)} stepwise resolution and \textbf{(b)} agentic dialogue. Each chart displays the success rate (purple bars, left axis), the total number of tokens used with standard deviation (black line, right axis), and the average tokens per request (blue dashed line, right axis). Insets provide a zoomed-in view of the average per-request token usage. Here, $N$ denotes the number of disks in the puzzle, and $p$ indicates the number of moves requested per substage.}

    \label{fig:HanoiResults}
\end{figure}

Fig.~\ref{fig:HanoiResults}\textit{a)} illustrates the results. The purple bars indicate the success rate across trials, while the two overlaid curves represent different aspects of token usage. The black curve shows the average total number of tokens consumed to solve the complete puzzle, calculated as the cumulative token count across all substages in a trial. In contrast, the blue dashed curve reflects the average number of tokens used per substage, capturing the typical cost of a single reasoning step within the larger sequence.

The inset in the top-right corner zooms into the lower range of the blue curve, allowing better visual comparison between configurations, as per-substage token usage remains relatively small in scale. This dual analysis reveals how overall reasoning effort and localized step efficiency evolve as task complexity increases.

There is a clear rise in problem complexity as the number of disks increases, reflected in the progressive drop in success rate. When the problem reaches higher levels of difficulty (i.e., $N \geq 8$), we observe a notable decrease in the total number of tokens used by the model during reasoning. However, this trend must be interpreted with caution, as the total token count (black line) is highly sensitive to the substage at which the model makes a mistake. If the LRM fails early in the resolution—e.g., during the first substage—the overall token consumption remains low, regardless of the overall complexity of the puzzle. Conversely, mistakes made in later substages accumulate more tokens before failure occurs.

For this reason, analyzing the average number of tokens per request (blue line) provides a more stable indicator of reasoning effort. Unlike total token usage, this metric is independent of the specific substage at which failure occurs and reflects only the intrinsic difficulty of each individual substage. Since substage complexity tends to grow with the overall task, this metric offers a clearer view of how demanding each reasoning segment is, independent of earlier or later errors in the full resolution.

The evolution of token usage, together with the success rate, confirms that when puzzle complexity increases - yet remains within the LRM’s solvable range - the token rate also rises accordingly, signaling increased effort to meet the challenge. However, when the puzzle exceeds the LRM’s capabilities, the token rate drops, reflecting early-stage failures and an implicit recognition of task infeasibility. This supports the interpretation that the complexity of each substage directly mirrors the global difficulty of the overall puzzle.

\subsubsection{Agentic Dialogue:} The procedure is equivalent to the previous experiment, evaluating the same configurations from $N=3$ to $N=10$, with 10 independent trials conducted for each (Fig.~\ref{fig:HanoiResults}~\textit{b}) presents the corresponding results.

Performance shows a clear drop compared to the stepwise resolution setup. The model begins to struggle as early as $N=4$, clearly exhibiting worse performance in solving the problem compared to the previous methodology. However, the token usage patterns reveal a contrasting trend. Unlike in the previous experiment—where total token usage decreases sharply as complexity increases and failure occurs early—in this case, the LRM agents continue to invest considerable effort even when they are ultimately unable to solve the task. Token rates remain high across configurations.

This behavior likely reflects that the agents recognize the task as solvable within their theoretical capacity, yet fail to discover the correct sequence of actions. This interpretation aligns with empirical observations: in most trials, the agentic system maintains extended dialogues without committing illegal moves, but ends up looping indefinitely—oscillating forward and backward—without ever converging to a valid solution.

\subsection{River Crossing}
\begin{figure}[ht]
    \centering
    \includegraphics[width=\linewidth]{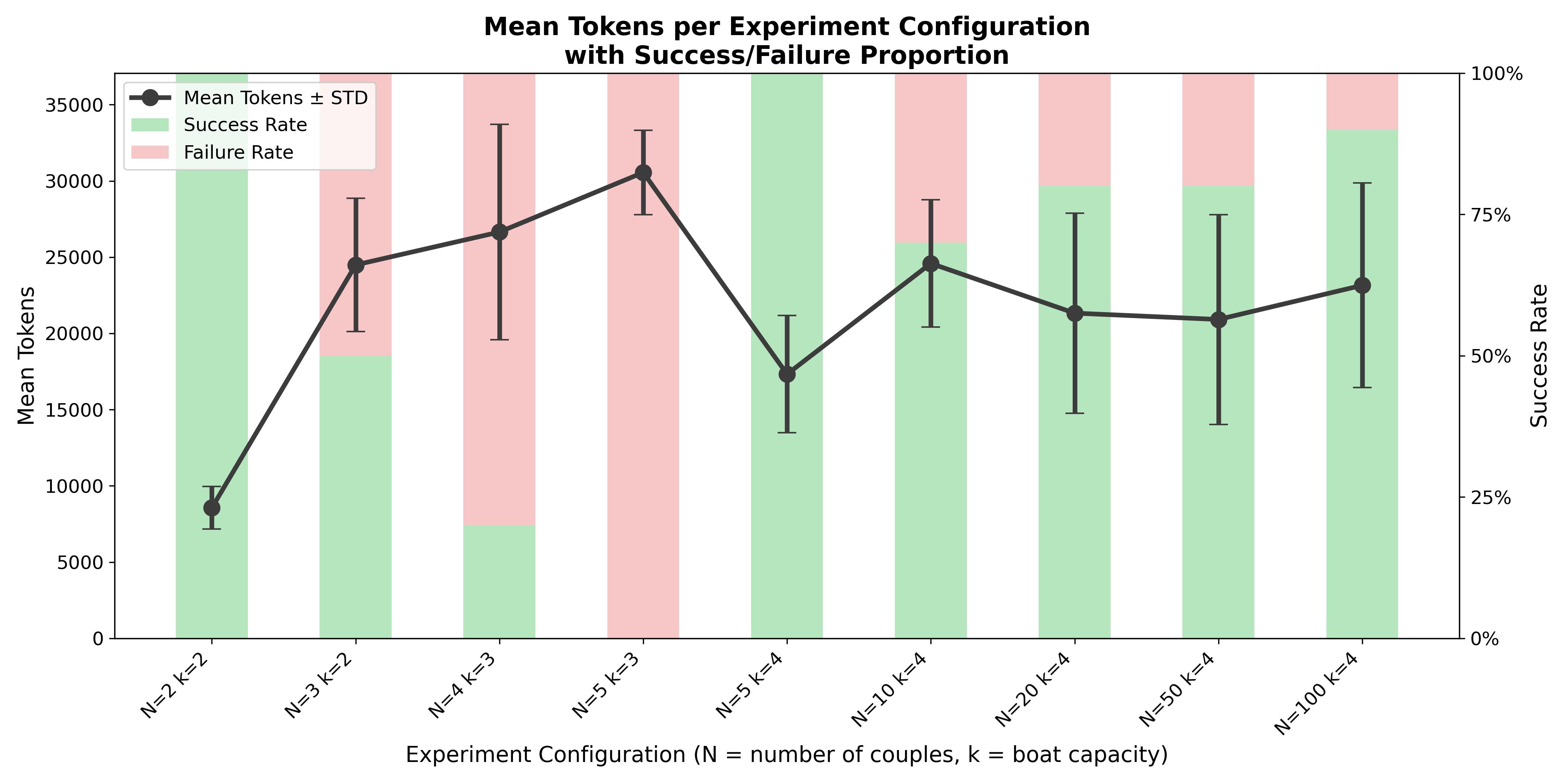}
    \caption{Performance of the LRM in the River Crossing puzzle across different solvable configurations, defined by the number of couples $N$ and boat capacity $k$. The black line with error bars represents the average number of tokens used by the model per configuration ($\pm$ standard deviation), while the colored bars indicate success (green) and failure (red) rates.}
    \label{fig:river_crossing_results}
\end{figure}

To test the capabilities of the LRM in solving this puzzle, we conduct a series of tests with varying complexity. The boat capacity is altered from $k = 2$ to $k = 4$, and the number of agent–actor couples ranges from $N = 2$ to $N = 100$, always respecting the constraints for solvable configurations. By restricting the experiments to valid instances only, we ensure that the performance reflects genuine reasoning ability, rather than limitations imposed by unsolvable setups. Each configuration is repeated 10 times (Fig.~\ref{fig:river_crossing_results}).

From Fig.~\ref{fig:river_crossing_results}, a clear trade-off between success rate and token consumption is observed. The most difficult configurations—$(N = 4,\, k = 3)$ and $(N = 5,\, k = 3)$—demand the highest number of tokens in the model’s attempt to solve them. Conversely, the simplest configurations—those with $k = 2$ and $k = 4$ regardless of $N$—pose little challenge to the system, which is able to solve them without significantly increasing token usage.

The results also show that increasing the puzzle size (i.e., increasing $N$) does not exhibit a direct correlation with reasoning complexity: neither the number of tokens consumed nor the success rates deteriorate consistently with larger $N$. Even in large-scale instances such as $(N = 100,\, k = 4)$—which require the model to perform up to 200 correct moves—the LRM solves the task reliably without significant effort or resource usage. This suggests that the challenge is not inherent to puzzle size itself.

Instead, the true difficulty appears to arise from specific configurations with a drastically reduced solution space. A particularly surprising effect emerges in configurations $(N = 4,\, k = 3)$ and, especially, $(N = 5,\, k = 3)$, where the system reaches a peak in resource demand while simultaneously showing the lowest efficiency.

\section{Discussion}

The experimental results presented in this paper provide a deeper insight into the reasoning abilities of Large Reasoning Models (LRMs) when solving puzzles of increasing complexity. Furthermore, they offer a structured framework to qualitatively compare our findings with those reported by Shojaee et al.~\cite{illusion-of-thinking}. 

\begin{figure*}[!ht]
\centering
\includegraphics[width=\textwidth]{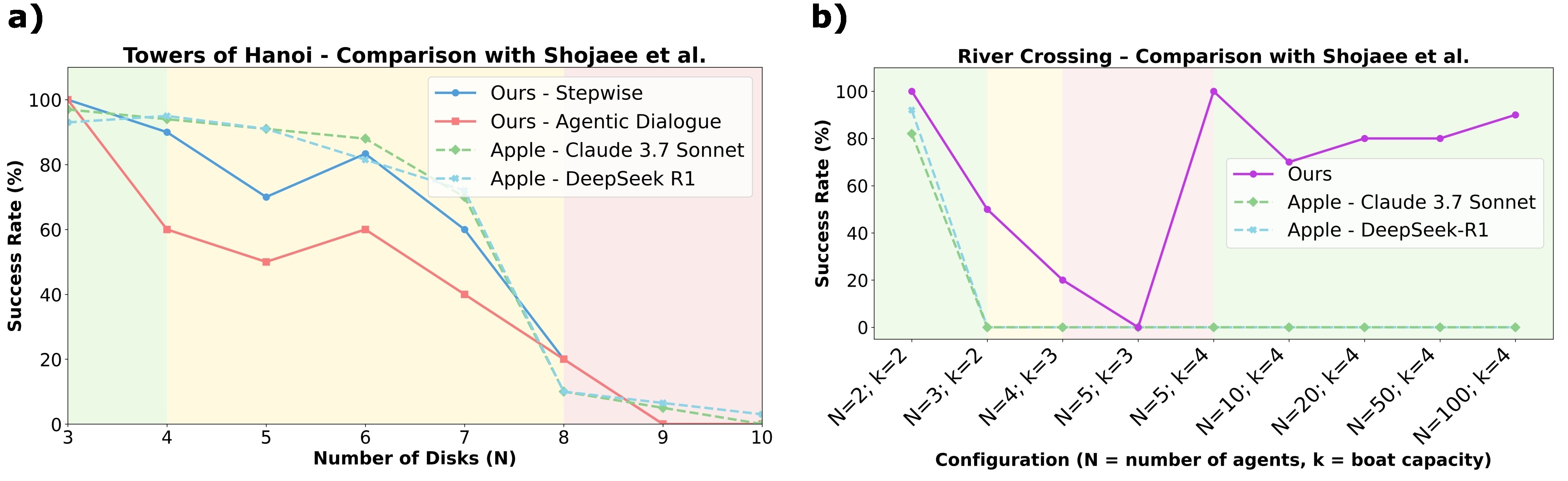}
\caption{Comparison of success rate trends between our experimental results and those reported by Shojaee et al.~\cite{illusion-of-thinking}. 
\textbf{(a)} Towers of Hanoi: stepwise prompting and agentic dialogue strategies are evaluated across increasing numbers of disks ($N$). 
\textbf{(b)} River Crossing: multiple configurations with varying number of agents ($N$) and boat capacities ($k$) are assessed. 
Apple's results (Claude 3.7 Sonnet and DeepSeek-R1) are estimated from the original paper, as exact success rate values are not publicly reported. 
This qualitative comparison highlights differences in reasoning dynamics and failure patterns, rather than direct performance benchmarking.}
\label{fig:comparission}
\end{figure*}

Fig.~\ref{fig:comparission} illustrates this comparative analysis. It contrasts the success rate trends from our experiments with those inferred from the original study by Shojaee et al., focusing on two distinct reasoning scenarios: the Towers of Hanoi (a) and the River Crossing problem (b). It is essential to clarify that the original paper does not disclose the exact success rate values for each configuration. Therefore, the performance curves for Claude 3.7 Sonnet and DeepSeek-R1 are manually estimated from the published plots and should be interpreted as approximations.

However, this limitation is not problematic for the purpose of our study. Our goal is not to benchmark models against each other in terms of absolute performance, but rather to analyze and contrast their general behavioral trends under varying levels of task complexity and prompting strategies. This comparative visualization reveals consistent patterns across both sources, and supports broader conclusions about the cognitive capabilities and limitations of LRM-based agents when faced with structured reasoning tasks.

Related to the Towers of Hanoi experiment (see Fig.~\ref{fig:comparission}\textit{a}), one of the most widely noted criticisms of the original study by Shojaee et al.~\cite{illusion-of-thinking} concerns the interpretation of failure on larger puzzles. While the authors attribute the drop in success rate exclusively to the increasing intrinsic complexity of the task, other voices in the research community suggest that these failures may also stem from technical limitations—namely~\cite{lawsen2025,khan2025}, the model’s awareness of its restricted context and output windows. According to this perspective, the model may possess a conceptual understanding of the solution, yet refrain from generating complete reasoning traces due to the anticipated length of the required output.

To examine these hypotheses, we implement two decomposition strategies—stepwise resolution and agentic dialogue. In both approaches, the original problem is split into smaller, independent sub-puzzles. Instead of prompting the LRM to solve the full puzzle in a single request, each prompt focuses on a localized task: continuing from the configuration achieved in the previous sub-puzzle and making a fixed number of valid moves toward the final solution.

Fig.~\ref{fig:comparission}\textit{a)} shows that, even when decomposing the task, performance does not significantly improve compared to the original study. Once puzzle complexity reaches approximately $N=8$, both strategies fail consistently. This result suggests that the difficulty lies not in the output length or window size, but in the structural complexity of the configuration itself. Therefore, beyond limitations in output space, current LRMs still exhibit intrinsic weaknesses when attempting to solve certain classes of complex reasoning problems.

The stepwise resolution strategy also reveals a clear correlation between problem complexity and the number of tokens consumed during reasoning. As the puzzle becomes more challenging, LRMs tend to use more tokens in their reasoning process. This trend holds as long as the model perceives the task as solvable. However, when the complexity surpasses what the LRM considers tractable, token usage drops sharply. In those cases, the model refrains from investing resources in a problem it implicitly deems unsolvable. This pattern is consistent with the observations reported by Shojaee et al.

Such dynamics suggest that high token expenditure may signal the model’s belief that a solution is within reach. Interestingly, the agentic dialogue approach offers an additional perspective on this behavior. Despite yielding the worst success rates in terms of accuracy, it exhibits a unique and revealing pattern. With two LRM agents interacting in alternating turns, the system does not “give up” on complex tasks. Instead, it shows the opposite trend: the more complex the puzzle, the greater the token consumption.

Unlike the stepwise or direct resolution approaches, where token usage drops when the model anticipates failure, the collaborating agents in agentic dialogue continue investing resources even when the task proves unsolvable. This persistent effort indicates that the agents internally estimate the problem as solvable, despite empirical evidence showing they are ultimately incapable of reaching the solution. They engage in extended exchanges filled with locally valid but unproductive moves—cycles that, while legal within the rules, never lead to resolution. This behavior reinforces the notion that such puzzles remain out of reach for current LRMs, not due to explicit violations of logic or game rules, but because of a deeper incapacity to structure and execute coherent long-horizon strategies.

Focusing on the River Crossing task (Fig.~\ref{fig:comparission}\textit{b}), we observe a substantial shift in performance when the problem is reconfigured relative to the original study. By restricting the experiments to initial configurations with known solvable solutions, the LRM demonstrates remarkably strong reasoning capabilities. What was previously its Achilles’ heel becomes one of its most reliable strengths, highlighting that an increase in problem size does not necessarily lead to reduced effectiveness. This finding reinforces the idea that LRM reasoning performance is highly dependent on the nature of the task itself and whether the model has been exposed to similar patterns during training. As a result, drawing general conclusions about their reasoning abilities proves challenging, as these systems exhibit strong context- and formulation-dependent behavior.

In the original study by Shojaee et al., all puzzles are intrinsically framed with a linearly increasing difficulty, proportional to problem size. However, unlike the Towers of Hanoi task, the River Crossing puzzle exhibits a non-linear difficulty profile. Our experiments reveal that the peak in difficulty occurs at the intermediate configuration \(N=5;\,k=3\), rather than at the largest instances. This behavior is characteristic of combinatorial problems, where the solution space is not uniformly challenging but instead divided into distinct regions~\cite{ref_kirkpatrick1994}.

Configurations with fewer agents, such as \(N<5;\,k=3\), have relatively few valid solutions, but those solutions are easy to discover due to their intuitive structure. On the other hand, configurations with more agents, such as \(N>5;\,k=3\), present a broader solution space. While they require longer action sequences, they do not necessarily demand the same level of precision, making it easier for the model to eventually find a valid path. The configuration \(N=5;\,k=3\), however, lies on the boundary between these two regimes. It contains very few viable solutions, each of which requires extremely precise reasoning and fine-grained planning. The absence of intuitive structural cues in this regime amplifies the cognitive load and makes it particularly prone to failure.

Such ``phase transition'' regions are notoriously difficult even for conventional search-based algorithms and are typically addressed using combinatorial optimization methods such as graph traversal, constraint satisfaction solvers, neural approximators, or evolutionary algorithms. In contrast, large language models operate through stochastic token generation, relying on learned semantic and syntactic patterns rather than systematic search mechanisms. This fundamental difference in approach makes them especially vulnerable in tightly constrained solution spaces, where even minor deviations can derail the entire reasoning process.

Looking ahead, the next steps in this project will primarily focus on scaling up the experimental framework. This includes running a larger number of tests per configuration to reduce statistical variance and expanding the set of configurations analyzed to capture a broader range of task complexities. Additionally, it would be valuable to assess whether the observed performance trends generalize across different LLM architectures. Comparing multiple models and tracking their progress over time could offer deeper insights into the evolution of reasoning capabilities as these systems continue to develop.

\section{Conclusion}
In this study, we present a replication and reconfiguration of the experiments from Apple’s \textit{The Illusion of Thinking}~\cite{illusion-of-thinking}, incorporating new ablation strategies such as stepwise resolution, agentic behavior, and configuration viability. Our aim is to contribute a fresh perspective to the ongoing debate surrounding the reasoning capabilities of large language models (LRMs).

Building on the theoretical positions of both proponents and critics of LRM reasoning, our findings reveal that the question is far from settled. We observe valid arguments on both sides, suggesting that reasoning capacity is difficult to generalize through benchmark tasks alone. Rather than a purely technical issue, this may ultimately be a philosophical or cognitive-level debate about what constitutes reasoning.

Our results support the notion that LRMs are capable of solving highly complex tasks requiring long, precise sequences of actions—such as River Crossing scenarios with more than 100 agent pairs—which would challenge even human solvers. However, as with any system attributed with reasoning ability, their competence varies by task. In our experiments, LRMs perform significantly better on River Crossing than on Towers of Hanoi.

Furthermore, we show that for LRMs, task difficulty does not correlate linearly with problem size. Certain intermediate configurations may prove nearly unsolvable, while larger instances are handled successfully. This non-linear profile underscores the need to evaluate LRM capabilities in terms of problem structure rather than size alone.

Finally, our findings—alongside those of Shojaee et al.—highlight a consistent link between task difficulty and token consumption during reasoning. LRMs tend to invest more tokens in problems they believe are solvable, but drastically reduce token usage when they perceive a task as beyond their capabilities. Thus, high token expenditure in difficult settings may indicate the system’s belief that a solution lies within reach, a pattern confirmed through collaborative agent setups.

Ultimately, using LRMs to solve these problems is equivalent to unleashing a stochastic search procedure, chain-of-thought sampling refined by reinforcement learning—inside a large, into discrete search space whose structure we have barely characterised. To make decisive progress, we must first map that terrain. The step-wise, agentic dissection introduced here—examining the space slice by slice (varying k, N, and puzzle size)—is a critical first step toward that goal\cite{rahwan2019machine}.\\

\subsubsection*{Code Availability.} The complete codebase and the prompts used in this study are available at: 
\url{https://github.com/11inaki11/Rethinking-The-Illusion-of-Thinking}

\subsubsection*{Acknowledgements}
The authors would like to thank José Hernández-Orallo for detailed comments on the analysis. IDV acknowledges funding from CSIC. MC is supported by the Horizon Europe Chips JU under the NexTArc CAR project (HORIZON-JU-Chips-2024-2-RIA) and by grant PID2023-150271NB-C21 from the Spanish Ministry of Science, Innovation, and Universities / Spanish State Research Agency (MICINN/AEI, DOI: 10.13039/501100011033). PRZ received a Training Program fellowship (PRE2020-634092049) from the Ministry of Science and Innovation. Additional funding was provided by Google.org through a grant awarded to Fundación General CSIC. Google.org had no role in the design, execution, analysis, reporting of this research, or in decisions related to publication or presentation.

%
%
%
%

\end{document}